\documentclass[journal]{IEEEtran}
\usepackage[utf8]{inputenc}
\usepackage{cite}
\usepackage{amsmath,amsthm,amssymb,amsfonts}
\usepackage{mathtools}
\usepackage{algorithmic}
\usepackage{graphicx}
\usepackage{textcomp}
\usepackage{subfiles}
\usepackage{enumitem}

\ifCLASSINFOpdf
\else
   \usepackage[dvips]{graphicx}
\fi
\usepackage{url}

\usepackage{graphicx}
\usepackage{booktabs}
\usepackage{floatrow}
\newfloatcommand{capbtabbox}{table}[][0.25\FBwidth]

\newcommand{\citep}{\cite}
\newcommand{\citet}{\cite}
\newcommand{\citeyear}{\cite}

\newtheorem{theorem}{Theorem}
\newtheorem*{theorem*}{Theorem}

\newtheorem*{lemma*}{Lemma}
\newtheorem*{remark*}{Remark}

\theoremstyle{definition}
\newtheorem{example}{Example}

\begin{document}

\title{On an $L^2$ norm for stationary ARMA processes}

\author{Anand Ganesh, Babhrubahan Bose, and Anand Rajagopalan
\thanks{Anand Ganesh, National Institute of Advanced Studies, IISc Campus, Bengaluru 560012, India (e-mail: anandg@nias.res.in).}
\thanks{Babhrubahan Bose, Department of Mathematics, Indian Institute of Science, Bengaluru,
560012 India (e-mail: babhrubahanbose@gmail.com).}
\thanks{Anand Rajagopalan, Cruise, San Francisco, USA (e-mail: anandbr@gmail.com)}
}

\markboth{Preprint, Anand Ganesh, Babhrubahan Bose, Anand Rajagopalan, 2026}
{Shell \MakeLowercase{\textit{et al.}}: Bare Demo of IEEEtran.cls for IEEE Journals}

\maketitle
\begin{abstract}

We propose an $L^2$ norm for stationary Autoregressive Moving Average (ARMA) models. We look at ARMA models within the Hilbert space of the past with present of a true purely linearly non-deterministic stationary process $X_t$, and compute the $L^2$ norm based on its Wold decomposition. As an application of this $L^2$ norm, we derive bounds on the mean square prediction error for AR(1) models of MA(1) processes, and verify these bounds empirically for sample data.
\newline
\newline
\end{abstract}


\IEEEpeerreviewmaketitle

\section{Introduction}

The current paper proposes an $L^2$ norm to study ARMA models and their relation to purely non-deterministic stationary processes. Similar norms for ARMA models have been proposed before, for the purpose for time series clustering \cite{Piccolo1990} or classification \cite{Martin2002}. But these treatments apply the norm to compare ARMA models with each other, but not to compare ARMA models with any other stationary process, this being sufficient for problems of clustering or classification.

Our paper revives the $L^2$ norm, briefly suggested by \cite{Piccolo1990}, but with an additional, more theoretical purpose. We seek a norm to compare ARMA models not only with each other, but also with purely linearly non-deterministic stationary processes more generally. In doing so, we would like to establish some notion of proximity between a model, and a true stationary process which may not be an ARMA process at all.

The idea of an ARMA model is historically tied to the idea of rational spectral density as in \cite{Doob1953, Whittle1983, Broersen2006}. These treatments identify a given process as being ARMA if their spectral density function, or equivalently the transfer function, is representible as a rational function. In contrast to these arguments, we have the heuristic argument of Box-Jenkins \cite{BoxJenkins1976} where they suggest that a given stationary process could have an ARMA model if the transfer function has a good rational approximation. A further pedagogical extension of this argument based on the Wold decomposition can be seen in \cite{Triacca2014}. These are attractive heuristic arguments, but it is unclear if they are defensible. Our discussion on innovation processes in Section \ref{sec_l2} leading up to Theorem \ref{hinfinity_sarason_theorem} is a careful attempt to make rigorous statements conveying a more limited but similar message.

The Box-Jenkins argument makes a distinction between a process and a model, an idea that has been articulated earlier by Whittle \cite{Whittle1963} and in occasional later work like \cite{Franke1985}. But this distinction is not a common one in literature. Most prediction theory starts with an ARMA process to make forecasts. Or, one estimates an ARMA model from given data based on some likelihood or least squares criterion, assuming that the data was generated by an ARMA process in the first place. Our paper keeps a clear distinction between the true process and a model, and suggests that an $L^2$ norm based on the Wold decomposition (or MA($\infty$) representation) provides a good way to measure their proximity. We connect these ideas in a triangle inequality (in Theorem \ref{triangle_inequality_theorem}) using the $L^2$ norm to bound the predictive error of a model.

As an application of these theoretical ideas, and also to verify them empirically, we look at modeling an MA(1) process with an AR(1) model, and derive some bounds on the prediction error (Table \ref{table:prediction_error}). The example also captures, in a simplified setting, the essential elements of the (undocumented) \verb|ar2arma| function \cite{Perktold2010} within the python \verb|statsmodels| package. The \verb|arima_process.ar2arma| function minimizes the integrated square error of the impulse responses while fitting an AR process to an ARMA model of a given order. This is equivalent to using an $L^2$ norm, though without naming it as such, and without clear justification.

Computerized model fitting algorithms tend to use maximum likelihood estimation as opposed to least squares techniques like Burg's algorithm which are faster, but considered less accurate. Our use of least square error and an $L^2$ norm may appear unusual in this context. Despite this, one justification for the $L^2$ norm is that ARMA models often get compared to various machine learning models including neural networks, and in this context, it is much easier to work with a simpler metric like the mean square error. Further, the likelihood estimation algorithms contain many assumptions like Gaussian priors, and suitable algorithms to initialize and run the Kalman filter. With all this complexity, it is often difficult to connect the error values obtained to the computed likelihood values. \cite{Wheeler2023} even notes that the inability to properly maximize model likelihoods can lead to misleading results where researchers may incorrectly believe that they are minimizing AIC or similar criteria. Given these reasons, it seems useful to look at mean squared error, and an associated $L^2$ norm.

It is a long standing problem to determine the order of an ARMA model, as for instance in \cite{Beguin1980, Liu1982, Gooijer1985, Choi1992, Broersen2006} and ongoing algorithmic work on automatic determination of the ARMA model \cite{HyndmanKhandakar2008, Wheeler2023}. We believe that the $L^2$ norm, or in essence the least squares distance, could be useful within the broader context of automatic model fitting algorithms, for instance, to initialize the ARMA model of a given order prior to running a maximum likelihood estimation algorithm.

Given the intuitive appeal of a simple $L^2$ norm, and the repeated appearance of such approximation ideas as in \cite{BoxJenkins1976, Piccolo1990, Martin2002, Perktold2010, Triacca2014}, we believe the problem merits further study.

\section{Literature Review}

An $L^2$ norm similar to ours, based on an $MA(\infty)$ representation, is mentioned briefly by Piccolo \cite{Piccolo1990}, but the idea is not pursued since it is found unsuitable for non-stationary processes. Piccolo goes on to use the AR($\infty$) coefficients instead, in a formal manner, to understand the distance between ARMA models, but he does not address issues around a common Hilbert space structure as we do in Section \ref{sec_l2}. While this formal approach suffices for model clustering, his notion of distance is not connected to properties of the underlying stationary process, for instance its variance as in Theorem \ref{triangle_inequality_theorem}. The use of AR($\infty$) coefficients could be problematic in other ways as well. For one, the past processes $X_{t-k}$ do not form an orthogonal basis, so the use of square error is not natural. Further, there is no guarantee of a unique AR($\infty$) decomposition so it is unclear if the distance is even well defined. Despite such potential objections, the formal approach seems to have worked for the clustering problems addressed by the paper.

More recently, Martin \cite{Martin2002} criticizes the possibility of an $L^2$ norm. He refers to the fact that there are multiple representations of a stationary process using MA, AR, and ARMA models, so picking a weighted $L^2$ distance based on just one representation may be misleading, and may not account for the poles and zeros of the system.  His criticism around uniqueness applies to Piccolo's work, but not to ours given the uniqueness of the Wold decomposition. Martin does not refer to the Wold decomposition at all, perhaps because of his focus on ARMA models in particular. As it happens, Martin's  transfer function $H(z)$ is itself a power series derived from the Wold decomposition.

\cite{Martin2002} goes on to define a cepstral norm based on the power spectrum function, an elegant approach that respects the multiplicative structure of rational functions, and the distribution of poles in the system. Our $L^2$ norm is complementary to Martin's cepstral norm. Where he looks at the logarithm of the spectrum, we look at the transfer function directly, as suggested by one of Martin's reviewers. To find the cepstral norm one starts with the logarithm of the power spectrum:
\begin{gather*}
\log P(z) = \log[x(z)x^*(\frac{1}{z})] = \sum_{n \in \mathbb{Z}} c_n z^{-n}
\end{gather*}

The $(c_n)$ are the cepstral coefficients, and the cepstral norm is based on a weighted $L^2$ distance on the cepstral coefficients $d(c, c') = \sum_{n=1}^{\infty} w_n |c_n - c_n'|^2$ for suitable weights $w_n$.

Related ideas on approximation have been touched upon by Hannan \cite{Hannan1980, Hannan1981} wherein he explores the idea of imposing a topology on ARMA models as a way to compare them with each other, and with the true process. But the approach is highly statistical and is closely tied to notions of consistent estimators.

\section{Preliminaries}

Let $(\Omega, \mathcal{F}, \mathbb{P})$ be a probability space. Let $L^2(\Omega)$ be the Hilbert space of random variables $X$ on $\Omega$ having finite second moment. Throughout the article, we shall be assuming that $X$ takes values in the set of complex numbers $\mathbb{C}$. Let us define a weakly stationary sequence of random variables $\{ X_n \}_{n=-\infty}^{\infty}$ on $\Omega$ to be a sequence in $L^2({\Omega})$ such that cov$(X_n, X_{n+k})$ only depends on $k$. Let us recall the celebrated Wold decomposition theorem for covariance stationary random processes.
\begin{theorem}
{\em Wold Decomposition} \citep[p.109]{Hamilton1994}: Any zero-mean covariance-stationary process $X_t$ can be represented in the form
\begin{align} \label{wold_decomposition_equation}
X_t = \sum_{j=0}^{\infty} x_j \epsilon_{t-j} + \kappa_t.
\end{align}
\end{theorem}
Here $\epsilon_t$ is white noise representing {\it linear forecasting error}, also called {\it innovation}, and $\sum_{j=0}^{\infty} x_j^2 < \infty$ (i.e. $\{x_j\} \in \ell^2$). Instead of the more standard assumption that $x_0 = 1$, we will assume, without loss of generality, that the $\epsilon_{t-j}$ are normalized. Now $\sum_{j=0}^{\infty} x_j \epsilon_{t-j}$ is the linearly non-deterministic component while $\kappa_t$ is called the lineary deterministic component. We focus on the case $\kappa_t = 0$ in which case $X_t$ is called a purely linearly non-deterministic process. It may be worth noting that the Wold decomposition theorem holds for (putatively) nonlinear and heterskedastic processes as long as they are stationary \cite{Scargle2020}. The normalization of $\epsilon_{t-j}$ allows us to compute the standard $L^2$ norm of $X_t$ in terms of $x_{j}$ as $\| X_t \|^2 = \sum_{j=0}^{\infty} x_j^2$.

Throughout this article, we shall be focusing on purely linearly non-deterministic processes, that is, random processes for which $\kappa_t = 0 ~ \forall~t~ \in ~ \mathbb{Z}$. The aim of this article is to establish theoretical justifications for Autoregressive Moving Average (ARMA) approximations of purely linearly non-deterministic covariance stationary processes. To that end, we are going to deviate from the typical order of beginning with a fixed stationary process $X_t$ and computing $\epsilon_t$ from $X_t$ and instead start with a sequence of innovations $\{\epsilon_j \}_{-\infty}^{\infty}$ and construct various stationary processes using \eqref{wold_decomposition_equation} (with $\kappa_t = 0$ as mentioned before).

The notation $\hat{X_t}$ refers to the optimal linear predictor for a given random variable $X_t$. Following standard procedure in prediction theory, we look at the space of past random variables as the span closure of past innovations. For a random variable $X_t$, the space of the past is $\overline{\text{sp}}\{ \epsilon_k\}_{k < t}$. The predictor $\hat{X_t}$ is often defined as $\hat{E}[X_t | X_{t-1}, X_{t-2}...]$, in other words, the expectation of $X_t$ given the past, and thus a random variable in its own right. This can also be regarded as a projection of $X_t$ on the space of past random variables. While there are procedural ways of understanding the predictor, say using the Levinson algorithm for ARMA models, this mathematical definition based on projections allows us compute norms and distances more easily as in the classic infimum results of Szego and Kolmogorov.

The prediction error associated with $\hat{X_t}$ is $\sigma(\hat{X_t} - X_t)$, the standard deviation associated with the difference between the prediction and the true value. Given that $\hat{X_t}$ is a random variable, we note that $\sigma(\hat{X_t} - X_t) = \|\hat{X_t} - X_t\|_2$.

\section{An $L^2$ Norm} \label{sec_l2}

Following Whittle \cite{Whittle1963}, we propose that an ARMA model $Y_t$ be viewed as an approximation of a purely linearly non-deterministic, covariance stationary random process $X_t$, and that the goal of ARMA model estimation is to find the optimal mean square approximation of $X_t$. We relate the mean square prediction error to the $L^2$ norm $\|Y_t - X_t\|_2$.

We associate with $X_t$ the space of its past including present $\hat{\mathcal{P}_t}$, that is the closed subspace spanned by $\{X_i | i \le t\}$ or equivalently $\{\epsilon_i | i \le t\}$ where $X_t = \sum_{j=0}^{\infty} x_j \epsilon_{t-j}$. This is a standard association where the space $\hat{\mathcal{P}_t}$ is variously denoted as $\mathfrak{M_t}$ by \cite{WienerMasani1957} or $H_t$ by \cite{Pourahmadi1992}. By constraining a model $Y_t$ to be in $\hat{\mathcal{P}_t}$, we can represent $X_t$ and $Y_t$ using a common set of basis vectors $\epsilon_i$, and measure their distance $\| X_t - Y_t\|_2$ based on the $\ell^2$ norm of their difference.

We now list out a series of definitions related to ARMA models. An ARMA$(m, n)$ model is typically associated with the following form:
\begin{gather} \label{eq:ARMA_epsilon}
    Y_t + \sum_{k=1}^{m} a_k Y_{t-k} = \epsilon_t + \sum_{k=1}^{n} b_k \epsilon_{t-k}.
\end{gather}

The processes $\epsilon_i$ above are intended to match the basis vectors of the Hilbert space $\mathcal{H}$ that we just described. This represents a substantial assumption, and in effects limits the scope of the ARMA model. We will now justify this limiting assumption in two steps. Suppose instead, that we had an ARMA model with a different sequence of innovations $\eta_i$ as follows:
\begin{gather} \label{eq:ARMA_eta}
    Y_t + \sum_{k=1}^{m} a_k Y_{t-k} = \eta_t + \sum_{k=1}^{n} b_k \eta_{t-k}.
\end{gather}

We will first assert that we can choose $\eta_i \in \mathcal{H}$ without loss of generality. If $\eta_i \notin \mathcal{H}$ we can still assume there is a common space like $L^2(\Omega)$, and thus consider the projections of $\eta_i$ onto $\mathcal{H}$, and along with that the projections of $Y_t$ on $\mathcal{H}$ as well. We call these projections $\eta_i'$ and $Y_t'$ respectively. While there is no guarantee that $\eta_i'$ continue to be orthonormal, it is clear that $\|Y_t' - X_t\| \le \|Y_t - X_t \|$, and thus $Y_t'$ is a better model than $Y_t$. It is possible that there are parsimonious models $Y_t'$ that involve non-orthogonal white noise processes $\eta_i'$. We will make another simplifying assumption by ignoring such models, or at least assuming that there exist equally good models $Y_t$ within $\mathcal{H}$ that have the same approximation error as $Y_t'$.

Assuming now that $\eta_i \in \mathcal{H}$, we can now represent $\eta_t$ in terms of $\epsilon_i$ since $(\epsilon_i)$ forms a basis of $\mathcal{H}$. Say $\eta_i = T \epsilon_i$. If we assume now, that the underlying model is time invariant with respect to the true process we can prove that $T$ is an $H^{\infty}$ functions of $L$ as follows:
\begin{theorem} \label{hinfinity_sarason_theorem}
Given a time invariant ARMA model with stationary process $Y_t \in \hat{\mathcal{P}_t}$, and suppose $Y_t$ has the innovation sequence $(\eta_t)$ such that $\eta_t = T \epsilon_t$ for some bounded operator $T$, we can assert that $T = \psi(L)$ where $\psi \in H^{\infty}$.
\end{theorem}
\begin{proof}
We start with a model $Y_t = \sum_{k=0}^{\infty} \alpha_k \eta_{t-k}$. We then have $Y_{t-1} = LY_t = \sum_{k=0}^{\infty} \alpha_k L \eta_{t-k}$. But then, by Wold decomposition we also have $Y_{t-1} = \sum_{k=0}^{\infty} \alpha_k \eta_{t- 1 - k}$. Equating these two, we have:
\begin{gather} \label{wold_ytm1}
    \sum_{k=0}^{\infty} \alpha_k L \eta_{t-k} = \sum_{k=0}^{\infty} \alpha_k \eta_{t- 1 - k}.
\end{gather}

Further,
\begin{gather*}
    \langle L \eta_k, L \eta_j \rangle = \langle L^* L \eta_k, \eta_j \rangle = \langle \eta_j, \eta_k \rangle.
\end{gather*}

Thus, $(L \eta_{t-k})_{k=0}^{\infty}$ is orthonormal since $(\eta_{t-k})_{k=0}^{\infty}$ is orthonormal. By uniqueness of Wold decomposition, and the two representations of $Y_{t-1}$ in \eqref{wold_ytm1}, we must have $L\eta_{t-j} = \eta_{t-j-1} ~\forall~ j \ge 0$. We thus have for all $k \ge 0$,
\begin{align*}
    L \eta_{t-k} &= \eta_{t- 1 - k} \\
    \implies LT \epsilon_{t-k} &=  T\epsilon_{t- 1 - k} \\
    \implies LT \epsilon_{t-k} &= TL\epsilon_{t- k}.
\end{align*}

That is, $LT \epsilon_{t-k} = TL \epsilon_{t-k} ~\forall~ k \ge 0$. Since $(\epsilon_{t-k})_{k \ge 0}$ is a basis, we have $LT = TL$, that is $T$ commutes with $L$. But then, by Sarason's interpolation theorem \cite{Sarason1967}, we have $T = \psi(L)$ for $\psi \in H^{\infty}$.

\end{proof}

For such a time invariant setup with $H^{\infty}$ function $\psi$, we have the following:
\[
    Y_t = \gamma(L) \eta_t = \gamma(L) \psi(L) \epsilon_t = \beta(L) \epsilon_t \approx \frac{p(L)}{q(L)} \epsilon_t.
\]

Thus, $q(L) Y_t = p(L) \epsilon_t$, which matches the suggested form \eqref{eq:ARMA_epsilon} above for monic polynomials $p(L)$ and $q(L)$, with original innovation sequence $\epsilon_t$.

Somewhat related to this, \cite{Hoon1996} makes some fine distinctions between innovation process and residual variances, but they don't attempt to impose a common Hilbert space structure, or to define a norm using this process. As further justification for the $L^2$ norm, and also as an application of the above Hilbert space ideas, we state the following theorem connecting the $L^2$ norm to the prediction error in a sort of bias-variance equation:
\begin{theorem} \label{triangle_inequality_theorem}
Given a purely non-deterministic stationary process $X_t$ and an ARMA model $Y_t \in \hat{\mathcal{P}_t}$, the total prediction error for $Y_t$ is bounded by the sum of the prediction variance of the model $Y_t$ and the approximation error between the model and the true process $\|Y_t - X_t\|_2$ as follows:
\begin{gather}
\sigma(\hat{Y_t} - X_t) \le \sigma(\hat{Y_t} - Y_t) + \|Y_t - X_t\|_2
\end{gather}
\end{theorem}
\begin{proof}
The prediction values $\hat{X_t}$ and $\hat{Y_t}$ are valid random variables. Using this,
\begin{align*}
\sigma(\hat{Y_t} - X_t) &= \|\hat{Y_t} - X_t\|_2 \\
                        &\le \|\hat{Y_t} - Y_t\|_2 + \|Y_t - X_t\|_2 \\
                        &= \sigma(\hat{Y_t} - Y_t) + \|Y_t - X_t\|_2.
\end{align*}
\end{proof}

\section{Model Identification} \label{sec_identification}

The following section represents a simplified variant of the \verb|statsmodels| \verb|ar2arma| function where we start with an MA($1$) process instead of a general AR($p$) process. As noted earlier, the formal validity of such norm calculations is subject to many restrictions on the innovations within the space of the past. But empirical behavior assuming Gaussian innovations suggests that the triangle inequality described in the Theorem \ref{triangle_inequality_theorem} could apply more generally in some approximate manner.

\begin{example} \label{example_0}
Consider the  following MA(1) stationary random variable $X_t$:
\[
    X_t = \epsilon_t + a \epsilon_{t-1}.
\]

and an AR(1) predictor $\hat{X_t} = \beta X_{t-1}$. Given this predictor we can associate with it the following AR(1) process within $\hat{\mathcal{P}_t}$, the past of $X_t$ as described earlier:
\[
    Y_t = \epsilon_t + \beta Y_{t-1}.
\]

We can represent $Y_t$ as follows:
\begin{gather*}
    (1 - \beta L) Y_t = \epsilon_t \\
    Y_t = \epsilon_t + \beta \epsilon_{t-1} + \beta^2 \epsilon_{t-2} \dots.
\end{gather*}

We can now compute $\sigma(\hat{Y_t} - X_t)$ as follows:
\begin{align*}
    \sigma(\hat{Y_t} - X_t) &= \sigma(\beta Y_{t-1} - X_t) \\
                            &= \sigma(\beta(\epsilon_{t-1} + \beta \epsilon_{t-2} + \beta^2 \epsilon_{t-1} \dots) \\
                            & \quad \quad \quad - (\epsilon_{t} + a\epsilon_{t-1})) \\
                            &= (1 + (\beta - a)^2 + \frac{\beta^4}{1 - \beta^2})^{\frac{1}{2}}.
\end{align*}

Now $\hat{Y_t} = \beta Y_{t-1}$ and $\| \hat{Y_t} - Y_t \|^2 = \sigma(\epsilon_t)^2 = 1$, and finally:
\[
    \|X_t - Y_t\| = ((\beta - a)^2 + \frac{\beta^4}{1 - \beta^2})^{\frac{1}{2}}.
\]

From these we get the following inequality:
\begin{gather*}
1 + ((\beta - a)^2 + \frac{\beta^4}{1 - \beta^2})^\frac{1}{2} \ge (1 + (\beta - a)^2 + \frac{\beta^4}{1 - \beta^2})^{\frac{1}{2}}.
\end{gather*}
This follows from Minkowski's inequality which implies $\sqrt{x} + \sqrt{y} \ge \sqrt{x+y}$ for $x, y \ge 0$, thus verifying Theorem \ref{triangle_inequality_theorem} for AR(1) models of MA(1) processes.

To obtain the optimal AR(1) predictor $\hat{X_t} = \beta X_{t-1}$ for $X_t$ we consider the following minimization problem:
\begin{gather*}
\min E | X_t - \beta X_{t-1}|^2
\end{gather*}
This is a standard mean square extremal used to obtain the ideal ARMA predictor for a given stationary process \cite{Whittle1963, Franke1985}. Let us find $\beta$ based on this minimization problem.
\begin{align*}
f(\beta) &= \min E | \epsilon_t + (\beta - a) \epsilon_{t-1} + \frac{\beta}{2} \epsilon_{t-2} |^2 \\
         &= \min 1 + (a^2 - 2a \beta + \beta^2) + a^2 \beta^2
\end{align*}

Taking derivatives with respect to $\beta$,
\begin{gather*}
f'(\beta) = -2a + 2\beta + 2a^2 \beta = 0 \implies \beta = \frac{a}{a^2 + 1}.
\end{gather*}

The above algebraic calculations were verified empirically for two specific MA(1) processes with $a=\frac{1}{2}$, and $a = \frac{-1}{2}$, and the results are shown in Table \ref{table:prediction_error}. We consider three AR(1) models. The first model is the optimal one with $\beta = \frac{a}{a^2 + 1} = \frac{2}{5}, \frac{-2}{5}$. The second model uses $\beta = a$, and we call it the Padé model, and the last model was obtained using the \verb|SARIMAX.fit()| method. The RMSE column represents measured RMS error for a one step prediction or $\| \hat{Y_t} - X_t\|$, and the last column named Bound lists the value of $1 + \| X_t - Y_t\|$. For the first row,
\begin{align*}
\| X_t - Y_t \|^2 &= (\frac{2}{5} - \frac{1}{2})^2 + (\frac{2}{5}) ^ 4 \dots = \frac{1}{100} + \frac{(\frac{2}{5})^4}{1 - (\frac{2}{5})^2} \\
                  &\approx 0.01 + 0.03 = 0.04.
\end{align*}

Thus $\| X_t - Y_t \| \approx 0.2$, and we can verify that $\| \hat{Y_t} - X_t\| \le \| \hat{Y_t} - Y_t \| + \| X_t - Y_t \| = 1.2$. The empirically measured root mean square error for this AR(1) model is $1.03$, and this is shown in the RMSE column of Table \ref{table:prediction_error}.

\end{example}
\begin{table}
\caption{Prediction Error Summary}
\label{table:prediction_error}
\begin{center}
\begin{tabular}{|c|c|c|c|c|}
 \toprule
 True MA(1) & AR(1) Predictor & Type & RMSE & Bound \\
 \midrule
  & $\hat{X_t} = \frac{2}{5}X_{t-1}$ & Opt & 1.03 &  1.2 \\
 $X_t = \epsilon_t + \frac{1}{2}\epsilon_{t-1}$ & $\hat{X_t} = \frac{1}{2}X_{t-1}$ & Padé & 1.04 & 1.29 \\
  & $\hat{X_t} = -0.4 X_{t-1}$ & Fit & 1.27 &  1.86 \\
\midrule
 & $\hat{X_t} = \frac{-2}{5}X_{t-1}$ & Opt & 0.98 & 1.2 \\
 $X_t = \epsilon_t - \frac{1}{2}\epsilon_{t-1}$ & $\hat{X_t} = \frac{-1}{2}X_{t-1}$ & Padé & 0.99 &  1.29 \\
 & $\hat{X_t} = 0.4 X_{t-1}$ & Fit & 1.33 &  1.86 \\
 \bottomrule
\end{tabular}
\end{center}
\end{table}

\section{Conclusion}

We have looked at the approximation error of ARMA models based on the $L^2$ norm of random variables. Our work takes Box and Jenkins' heuristic transfer function approximation argument, and attempts to create a rigorous framework around it. Given the intuitive appeal, and the repeated appearance of such approximation ideas as in \cite{BoxJenkins1976, Piccolo1990, Martin2002, Perktold2010, Triacca2014}, it would be helpful to study this problem more thoroughly.

We provide a careful definition of ARMA models within the space of the past, wherein the ARMA model shares a Hilbert space structure with the true process. This leads to Theorem \ref{hinfinity_sarason_theorem} on time invariant models, and an $L^2$ norm used to bound the prediction error as in Theorem \ref{triangle_inequality_theorem}.

We find the optimal AR(1) model for an MA(1) process in an algorithmic manner using our proposed $L^2$ norm, verifying Theorem \ref{triangle_inequality_theorem} both algebraically, as well as empirically based on software simulation. Our empirical verification is currently limited to MA(1) processes with Gaussian innovations and AR(1) models. We would like to generalize our results to ARMA models more generally.

\section*{Acknowledgment}
Our sincere thanks to Prof. Nithin Nagaraj, Prof. Lakshminarayanan Subramanian, Dr. Shankar Viswanathan, Dr. Ajay Shenoy and Dr. Srikanth Pai for helpful discussions.
\bibliographystyle{IEEEtran}
\bibliography{IEEEabrv,ieee}

\end{document}